\documentclass[letterpaper]{article}
\usepackage{aaai}
\usepackage{times}
\usepackage{helvet}
\usepackage{courier}
\usepackage{natbib}
\usepackage{multicol}
\usepackage{multirow}
\usepackage{booktabs}
\usepackage[table]{xcolor}
\usepackage{bibentry}
\usepackage{graphicx}
\usepackage{algorithm}
\usepackage{amsmath}
\usepackage{amssymb}
\usepackage{subcaption}
\usepackage{subfig}
\begin{document}
%
\title{Rad-JEPA 3D: Radiology Joint-Embedding Predictive Model \\ for 3D Computed Tomography}
\author{
  Quoc-Huy Trinh\textsuperscript{1} \and
  Minh-Van Nguyen\textsuperscript{2} \and
  Ulas Bagci\textsuperscript{1} \\
  \textsuperscript{1}Northwestern University \quad
  \textsuperscript{2}Technical University of Denmark
}
\maketitle
\begin{abstract}
Self-supervised pretraining is central to 3D medical image analysis, where
unlabeled CT volumes are abundant but expert annotations are scarce. Yet existing
volumetric encoders often fail to preserve the coarse spatial and geometric
structure that downstream reasoning depends on, limiting their performance on
organ disentanglement, abnormality detection, and spatial understanding when
paired with language models. We introduce Rad-JEPA~3D, a joint-embedding
predictive framework that learns volumetric CT representations by predicting the
latent features of a complete scan from a masked view. At its core is a hybrid
H-Mamba encoder that fuses a Mamba state-space branch, which models inter-slice
continuity through sequential scanning, with a grouped-query attention branch,
which captures cross-plane spatial context, combined through a lightweight
per-token router. To improve the quality of intermediate representations, we further propose Hidden States Orthogonal Regularization (HSOR), which aligns student–teacher hidden states and reduces feature redundancy throughout the encoder. This layer-wise regularization produces more consistent and discriminative volumetric representations, leading to improved performance on organ recognition and spatial reasoning tasks.
Pretrained on approximately 120{,}000 CT scans, Rad-JEPA~3D attains
state-of-the-art results despite its compact size: with only $4.0$B total
parameters, it
achieves competitive results with state-of-the-art on closed-ended VQA and the best average
spatial-reasoning score on the Spatial-Med benchmark. Ablation studies confirm
that the hybrid block and HSOR contribute complementary gains, and that the
induced spatial structure can substitute for raw language-model scale on
volumetric reasoning tasks. URL: https://huyquoctrinh.github.io/radjepa3d/.
\end{abstract}

\section{Introduction}

Learning transferable visual representations from unlabeled data has become a
cornerstone of medical image analysis, where expert annotations are costly and
scarce while unlabeled scans are abundant. Self-supervised learning (SSL)
addresses this by pretraining encoders on label-free pretext objectives:
joint-embedding and contrastive methods that align augmented views
\citep{chen2020simple, he2020momentum, grill2020bootstrap}, masked image
modeling that reconstructs corrupted regions \citep{he2022masked}, and
self-distillation approaches such as DINO and DINOv2 that match teacher--student
distributions without negatives \citep{caron2021emerging, oquab2023dinov2}.
These recipes yield features that rival supervised pretraining and have given
rise to domain-specific foundation models, including RAD-DINO for chest
radiography \citep{perezgarcia2024raddino} and UNI for histopathology
\citep{chen2024uni}. Because CT and MRI are inherently volumetric, a parallel
line of work extends SSL to 3D, from restoration-based Models Genesis
\citep{zhou2021models} and multi-objective Swin~UNETR pretraining
\citep{tang2022self} to large-scale volume-contrastive frameworks such as VoCo
\citep{wu2024voco}, in order to capture the cross-plane anatomical context that
slice-wise 2D processing discards.

Vision-Language Models (VLMs), such as LLaVA~\citep{llava}, Qwen2-VL~\citep{qwen2vl}, and InternVL~\citep{internvl}, have recently demonstrated strong capabilities in joint visual-textual understanding across natural-image domains. This progress has rapidly extended into biomedicine, where LLaVA-Med~\citep{llavamed}, UniMed-VL~\citep{unimedvl}, and HealthGPT~\citep{lin2025healthgptmedicallargevisionlanguage} have achieved early success on diagnostic tasks over 2D images. More recently, M3D~\citep{bai2024m3d}, BTB3D~\citep{hamamcibetter}, and Med3D-VLM~\citep{xin2025med3dvlm} have taken the first steps toward integrating 3D volumetric representations into large language models for volumetric reasoning. However, as highlighted by SpatialMed~\citep{trinh2026beyond} and CT-SpatialVQA~\citep{monon2026lost}, these models underutilize the rich spatial information inherent in 3D volumetric data, which limits their performance on spatial understanding, organ disentanglement, and abnormality detection. We attribute this limitation to the absence of coarse spatial structure in the dense features produced by the volumetric encoder, which makes it difficult for the language model to recover and reason over such information.

Recently, V-JEPA~\citep{bardes2023v}, V-JEPA 2~\citep{assran2025v}, and V-JEPA 2.1~\citep{mur2026v} have proposed self-supervised frameworks that learn world representations directly from video. Their effectiveness stems largely from the architecture's ability to capture temporal structure across long sequences of video frames. Motivated by this observation, we propose \textbf{Rad-JEPA 3D}, a radiology joint-embedding predictive model for 3D Computed Tomography. We hypothesize that, by encoding the sequence of small cubes that compose a volume, the model can inherently represent coarse spatial structure alongside the dense features extracted from the volumetric data. In addition, we observe that intermediate encoder representations can become increasingly redundant across depth, limiting their discriminability for organ-level and spatial reasoning tasks. To address this, we propose Hidden States Orthogonal Regularization \textbf{(HSOR)}, which orthogonally regularizes the set of hidden states across encoder layers, encourages cross-view consistency and feature diversity throughout the network, helping spatially relevant information remain accessible to downstream tasks.

In summary, our contributions are threefold:
\begin{itemize}
    \item We introduce \textbf{Rad-JEPA 3D}, an efficient self-supervised architecture based on state-space models that captures dense representations of 3D Computed Tomography encoding both coarse spatial and geometric information from the CT scan. The model is trained on approximately 120{,}000 CT scans.
    \item We propose HSOR, a layer-wise regularization objective that aligns student and teacher representations while reducing redundancy in hidden features and encoder weights. Experiments show that this regularization improves spatially demanding downstream tasks when combined with the hybrid encoder.
    \item We conduct extensive experiments to evaluate the effectiveness of our SSL architecture across spatial question answering, visual question answering, and classification.
\end{itemize}

\section{Related Works}

\paragraph{Self-Supervised Learning in Medical Imaging.} In medical imaging, particularly in Computed Tomography (CT), the lack of expert annotation for a large number of
CT scans have made self-supervised learning (SSL) an attractive
paradigm for learning transferable representations without manual labels.
The initial works on 2D such as \citep{chen2020simple, he2020momentum,
grill2020bootstrap} adapt contrastive and joint-embedding methods that align augmented. Following that, several works such as MICLe~\citep{azizi2021big}, UNI~\citep{chen2024uni}, Virchow \citep{vorontsov2024virchow}, and Rad-Dino~\citep{perezgarcia2024raddino} are then leveraged the masked image modeling and the self-distillation approaches following DINO~\citep{oquab2023dinov2,simeoni2025dinov3} to let the model learn the representation for the 2D medical downstream tasks. 
Although the strong representation of the visual embedding from previous works, the 3D CT-scan is inherently three-dimensional, and the information of anatomical structures spans multiple slices; cross-plane
Context and inter-slice continuity that are critical for tasks such as organ/lesion diagnostic reasoning, and spatial reasoning.

To deal with this problem, recently, several works such as DINO3~\citep{xu20253dino}, MST~\citep{muller2025medical}, and VolTA3D~\citep{makawana2026volta} proposed the adaptation of the Masked Image Modeling by masking the random surface of the medical imaging. However, there are several limitations to these works. Firstly, they can not learn the structure or the cross-plane information across the CT slices due to the masking on the main slice only. Secondly, the adaptation based on the 3D Vision Transformer make the model be difficult to capture spatial and inter-slice continuity due to the need for the large-scale transformer architecture to capture this information. 

To address the challenges from Self-Supervised representation in medical imaging, particularly with 3D data, we leverage the hybrid architecture in Rad-JEPA 3D by leveraging the hybrid model between the mamba and transformer branches to capture continuity information across CT cubes and spatial information from the CT volumes, thus create the benefit of the model for visual question answering and spatial question answering.

\paragraph{Joint Predictive Modeling in 3D Medical Imaging.}
Different from masked image modeling, which reconstructs pixels, JEPA predicts the latent representation of masked regions from their surrounding context, encouraging high-level semantic learning~\citep{assran2023ijepa}. This paradigm has been extended to video through spatiotemporal prediction~\citep{bardes2024vjepa,assran2025vjepa2} and to medical imaging for brain dynamics, ultrasound, and multimodal CT--EHR learning~\citep{dong2024brainjepa,radhachandran2026usjepa,li2025selfsupervised}. However, existing medical JEPA methods remain limited in modeling long-range inter-slice and cross-plane dependencies and may exhibit weak downstream transfer or external generalization~\citep{ergun2026masked,li2025selfsupervised}. For this reason, Rad-JEPA 3D is proposed to address these limitations with a hybrid Mamba--Transformer encoder for volumetric dependency modeling and Hidden States Orthogonal Regularization (HSOR) to improve cross-view consistency and reduce feature redundancy.


\begin{figure*}[t]
    \centering
    \includegraphics[width=0.95\textwidth]{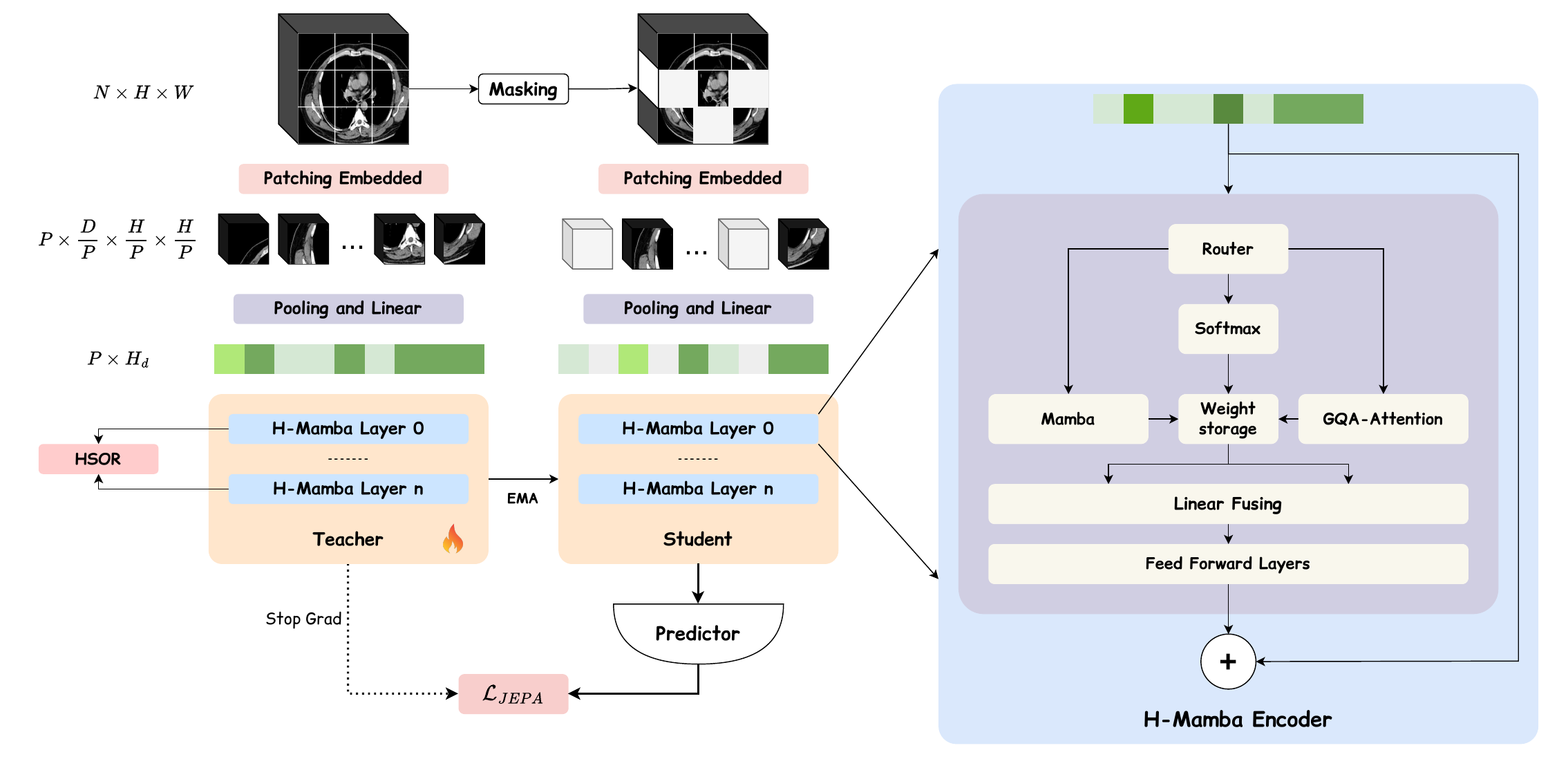}
    \caption{Overall architecture of Rad-JEPA~3D. A CT volume is tokenized into cubes and passed to two weight-sharing encoders: a student that sees a masked view and an EMA teacher (stop-gradient) that sees the full volume. A predictor aligns student features to the teacher's at the masked positions via the JEPA loss. Each encoder stacks H-Mamba layers (right), which fuse a Mamba and GQA branch through a per-token router; HSOR regularizes the student's per-layer hidden states.}
    \label{fig:problem_statement}
    \vspace{-3mm}
\end{figure*}
\section{Method}
\label{sec:method}

Rad-JEPA~3D, shown in Figure~\ref{fig:problem_statement}, learns volumetric CT
representations by predicting latent features of a complete scan from a masked
view. A student encoder $f_{\text{student}}$ processes the masked volume, while
an EMA teacher $f_{\text{teacher}}$ processes the intact volume with
stop-gradient. An MLP predictor maps student features to the teacher space for
alignment at masked positions, together with student-side regularization. The
encoder stacks H-Mamba blocks that combine Mamba-2~\citep{mamba2} for
inter-slice continuity and GQA~\citep{ainslie2023gqa} for long-range spatial
context through a lightweight router.

\subsection{Cube representation embedding}
\label{sec:input}
To transform the volumetric data into the representation for the coarse modeling,
the Cube Tokenizer is proposed so that the encoder can predict the coarse spatial
structure across the volume rather than processing slices in isolation. Let the
input be $\mathbf{X}\in\mathbb{R}^{C\times D\times H\times W}$, where $C{=}1$ for
CT intensity and $D$, $H$, $W$ are the number of axial slices, height, and width.
We partition $\mathbf{X}$ into cubes of size $p=(p_D,p_H,p_W)$ and
embed each into a $d$-dimensional token with a single strided 3D convolution whose
kernel and stride both equal $p$, so receptive fields tile the volume without
overlap:
\begin{equation}
\begin{gathered}
    \mathbf{Z} = \mathrm{Flatten}\big(\mathrm{Conv3D}(\mathbf{X})\big) + \mathbf{E}_{\mathrm{pos}}
    \;\in\;\mathbb{R}^{P\times d}, \\[4pt]
    \text{with }P=\tfrac{DHW}{p_D p_H p_W}.
\end{gathered}
\end{equation}
Tokens are flattened in raster order, and the learnable positional embedding
$\mathbf{E}_{\mathrm{pos}}\in\mathbb{R}^{P\times d}$ restores the 3D arrangement
discarded by flattening, preserving the spatial ordering that the encoder relies
on. The resulting sequence $\mathbf{Z}$ is consumed by the context and target
encoders.

\subsection{H-Mamba Encoder}
\label{sec:hmamba}
To capture both coarse spatial structure and the dense feature of the volumetric
data, each H-Mamba layer is constructed by mixing tokens with two complementary
operators, a Mamba branch and a grouped-query attention (GQA) branch. Their
feature outputs are then combined through a lightweight router, with a residual
connection around the block. Let $\mathbf{Z}\in\mathbb{R}^{P\times d}$ be the
layer input and $\tilde{\mathbf{Z}}=\mathrm{LayerNorm}(\mathbf{Z})$.

\paragraph{Branches.}
The Mamba branch is a selective state-space model~\citep{mamba2} that scans the
sequence with input-dependent parameters, giving linear-time modeling of the
long-range dependencies that arise from 3D volumes. The GQA branch shares
key/value projections across groups of query heads, retaining content-based,
all-to-all retrieval at reduced memory and compute relative to full multi-head
attention; rotary position embeddings~\citep{su2024roformer} inject the $(i_D,i_H,i_W)$ grid
coordinate of each cube into attention. Both branches process the full normalized
sequence in parallel, producing $\mathrm{Mamba}(\tilde{\mathbf{Z}})$ and
$\mathrm{GQA}(\tilde{\mathbf{Z}})$.

\paragraph{Routed fusion.}
To assign each token to the branch best suited to it, a lightweight linear router
is proposed. Router logits are mapped to branch probabilities
$\mathbf{g}=\mathrm{softmax}(\tilde{\mathbf{Z}}\,\mathbf{W}_r)\in\mathbb{R}^{P\times 2}$,
$\mathbf{W}_r\in\mathbb{R}^{d\times 2}$, and each token selects its higher-scoring
branch, weighted by that branch's confidence:
\begin{equation}
  b_i=\arg\max_{j}\,g_{i,j},\qquad
  \mathbf{h}_i=g_{i,b_i}\cdot
  \begin{cases}
    \mathrm{Mamba}({\mathbf{Z}})_i, & b_i=1,\\
    \mathrm{GQA}(\tilde{\mathbf{Z}})_i, & b_i=2.
  \end{cases}
\end{equation}
The block output is residual: $\mathbf{Z}'=\mathbf{Z}+\mathbf{h}$. To prevent the
router from collapsing onto a single branch, we add a load-balancing regularizer
$\mathcal{L}_{\mathrm{bal}}$ that encourages an even assignment across branches,
weighted by $\lambda_{\mathrm{bal}}$ in the total loss.

\paragraph{Router load balancing.}
To prevent all tokens from being assigned to the same branch, we encourage the
average routing distribution within each batch to be uniform. Let
\begin{equation}
\bar{g}_j
=
\frac{1}{BP}
\sum_{b=1}^{B}
\sum_{i=1}^{P}
g_{b,i,j}
\end{equation}
denote the average probability assigned to branch $j$. The load-balancing loss is
\begin{equation}
\mathcal{L}_{\mathrm{bal}}
=
\sum_{j=1}^{2}
\left(
\bar{g}_j-\frac{1}{2}
\right)^2.
\end{equation}

\subsection{Hidden States Orthogonal Regularization (HSOR)}
\label{sec:hor}
The masked-prediction objective primarily supervises the final encoder output, leaving middle representations weakly constrained. Consequently, hidden dimensions can be redundant or inconsistent between the masked student view and the complete teacher view. To address this limitation, we introduce HSOR to regularize representations throughout the middle representation using layer-wise student–teacher alignment and soft weight orthogonality.

\paragraph{Cross-correlation alignment.}
Let $\mathbf{s}^{(\ell)},\mathbf{t}^{(\ell)}\in\mathbb{R}^{B\times d}$ denote the
mean-pooled hidden states of the student $f_{\text{student}}$ (masked view) and
teacher $f_{\text{teacher}}$ (full view) at layer $\ell$, batch-standardized along
the feature dimension. We form their cross-correlation matrix
$\mathbf{C}^{(\ell)}=\tfrac{1}{B}\,\mathbf{s}^{(\ell)\top}\mathbf{t}^{(\ell)}\in\mathbb{R}^{d\times d}$
and drive it toward the identity, following the redundancy-reduction principle of
Barlow Twins~\citep{zbontar2021barlow}: on-diagonal entries are pulled to $1$ so
that the corresponding student and teacher features agree, while off-diagonal
entries are pushed to $0$ so that distinct feature dimensions remain decorrelated,
\begin{equation}
  \mathcal{L}_{\mathrm{align}}=\frac{1}{L}\sum_{\ell=1}^{L}
  \sum_{k}\big(C^{(\ell)}_{kk}-1\big)^2
  + \lambda_{\mathrm{off}}\sum_{k\neq k'}\big(C^{(\ell)}_{kk'}\big)^2 .
\end{equation}
Applying this objective at every layer encourages cross-view consistency and discourages redundant features throughout the encoder, rather than constraining only its final output.

\paragraph{Weight orthogonality.}
To further discourage redundant, collapsed filters, we apply a soft orthogonality
penalty to the student encoder's linear and convolutional weight matrices
$\mathbf{W}\in\mathcal{W}$. Each $\mathbf{W}$ is flattened to a 2D matrix
$\mathbf{W}\in\mathbb{R}^{m\times n}$ (output channels $\times$ fan-in). Since a
matrix admits at most $\min(m,n)$ orthonormal vectors, we penalize the smaller of
the two Gram matrices,
\begin{align}
  \mathbf{G}(\mathbf{W}) &= \begin{cases} \mathbf{W}\mathbf{W}^{\top} & \text{if } m \le n,\\ \mathbf{W}^{\top}\mathbf{W} & \text{otherwise,} \end{cases} \\
  \mathcal{L}_{\mathrm{reg}} &= \frac{1}{|\mathcal{W}|}\sum_{\mathbf{W}\in\mathcal{W}} \big\lVert \mathbf{G}(\mathbf{W})-\mathbf{I}\big\rVert_2.
\end{align}
Taking $\mathbf{W}^{\top}\mathbf{W}$ unconditionally would be ill-posed for wide
matrices ($m<n$), where $\operatorname{rank}(\mathbf{W}^{\top}\mathbf{W})\le m<n$
and the penalty is bounded below by $n-m$.

\subsection{Latent Predictor}
\label{sec:predictor}

Given the masked token sequence $\mathbf{Z}_{\mathcal{M}}$, the student
encoder produces contextual representations
$f_{\text{student}}(\mathbf{Z}_{\mathcal{M}})$. A four-layer MLP
predictor $g_{\psi}$ with hidden dimension 192 maps these representations
to the teacher embedding space. For each masked position
$i\in\mathcal{M}$, the predicted representation is
\begin{equation}
    \widehat{\mathbf{t}}_{i}
    =
    \left[
    g_{\psi}\!\left(
    f_{\text{student}}(\mathbf{Z}_{\mathcal{M}})
    \right)
    \right]_{i}.
\end{equation}
The predictor is shared across all masked positions, and
$\widehat{\mathbf{t}}_{i}$ is aligned with the corresponding
stop-gradient teacher target
$\left[\operatorname{sg}\!\left(
f_{\text{teacher}}(\mathbf{Z})
\right)\right]_{i}$ through the JEPA objective.

\subsection{Masked Denoising in Representation Space}
\label{sec:masked}
Pixel-space reconstruction forces the model to reproduce low-level CT texture and
noise that carry little anatomical meaning. To avoid this, Rad-JEPA~3D predicts
the masked content in \emph{representation space}: given a volume masked at the
positions $\mathcal{M}$, the model predicts the target encoder's features at those
positions rather than their voxel intensities. The student encoder
$f_{\text{student}}$ processes the masked token sequence $\mathbf{Z}_{\mathcal{M}}$,
a predictor $g_\psi$ maps its output into the target space, and the EMA teacher
$f_{\text{teacher}}$ processes the full sequence $\mathbf{Z}$ to supply the
prediction targets. Both are trained jointly by minimizing the $\ell_1$ distance
at the masked positions,
\begin{equation}
  \mathcal{L}_{\mathrm{JEPA}}
  =\frac{1}{|\mathcal{M}|}\sum_{i\in\mathcal{M}}
  \Big\lVert\, \widehat{\mathbf{t}}_{i}
  -\mathrm{sg}\big[f_{\text{teacher}}(\mathbf{Z})\big]_i \,\Big\rVert_1 ,
  \label{equa:masked_training}
\end{equation}
where $\mathrm{sg}(\cdot)$ is the stop-gradient and the teacher parameters
$\theta_{\text{teacher}}$ track the student parameters $\theta_{\text{student}}$
by exponential moving average,
$\theta_{\text{teacher}}\leftarrow\tau\,\theta_{\text{teacher}}+(1-\tau)\,\theta_{\text{student}}$.
Because the target is a learned representation rather than raw intensity, the
objective concentrates capacity on semantic and spatial structure instead of
reconstructing scanner noise.

\subsection{Training Objective}
\label{sec:objective}
The encoder and predictor are trained end-to-end by combining the masked
representation loss with the two HSOR terms and the router load-balancing term
$\mathcal{L}_{\mathrm{bal}}$, which discourages the per-token router from collapsing
onto a single branch:
\begin{equation}
  \mathcal{L}=\mathcal{L}_{\mathrm{JEPA}}
  +\gamma_{\mathrm{align}}\,\mathcal{L}_{\mathrm{align}}
  +\gamma_{\mathrm{reg}}\,\mathcal{L}_{\mathrm{reg}}
  +\lambda_{\mathrm{bal}}\,\mathcal{L}_{\mathrm{bal}} .
\end{equation}
After carefully tuning, we set $\gamma_{\mathrm{align}}{=}0.1$, $\gamma_{\mathrm{reg}}{=}10^{-6}$,
$\lambda_{\mathrm{bal}}{=}0.005$, and $\lambda_{\mathrm{off}}{=}0.005$ in all
experiments.
\section{Experimental Setup}

\subsection{Implementation Details}
\label{sec:implementation}

Rad-JEPA~3D uses a representation dimension of 384, with eight
Mamba-2 layers and six GQA-based Transformer layers. The visual encoder
contains approximately 19M parameters. Each CT volume is resized to
$128\times256\times256$ and tokenized into non-overlapping
$8\times8\times8$ cubes. During self-supervised pretraining, we randomly
mask $75\%$ of the cube tokens. A four-layer MLP predictor with a hidden
dimension of 192 maps the student representations into the teacher
embedding space. We train Rad-JEPA~3D for 10 epochs on eight NVIDIA A6000 GPUs using AdamW~\citep{adamw} with an initial learning rate of $10^{-4}$ and a cosine learning-rate
schedule. The local batch size is 32 per GPU, resulting in a global batch
size of 256. The teacher parameters are updated using an exponential
moving average of the student parameters. We select the checkpoint
according to the validation JEPA loss.

For downstream vision--language training, the pretrained visual encoder
is connected to either Qwen2.5-3B or Qwen3-4B, producing complete models
with approximately 3.0B and 4.0B parameters, respectively. We follow the
LLaVA visual-instruction-tuning procedure and apply LoRA to reduce the
number of trainable language-model parameters. Additional architectural,
optimization, and decoding configurations are provided in the
supplementary.

\subsection{Dataset}

Rad-JEPA 3D is trained on the publicly available M3D dataset~\citep{bai2024m3d},
collected with informed consent and ethical approval by the original work. All data were de-identified, and
no additional consent or approval was required for this
secondary analysis. For the pretraining stage, we use approximately 120,000 unique CT volumes from M3D-Cap, and M3D-VQA provides approximately 662,000 instruction-response pairs associated with these volumes.
Following this, we conduct post-training for the diagnostic reasoning and spatial reasoning tasks by using the text annotations from M3D-CAP and M3D-VQA training sets. All image volumes were resampled to a fixed size of 128x256×256. We divide the CT volumes into patient-disjoint training,
validation, and test sets containing 115,000, 3,000, and 2,000
volumes, respectively. No validation or test patient is included
during self-supervised pretraining or instruction tuning. For the evaluation set, we evaluate medical image visual question answering on both closed-loop and open-loop M3D-VQA test sets and spatial reasoning on the Multiple choice questions benchmark from SpatialMed~\citep{trinh2026beyond}.


\section{Results}
\begin{figure}[!ht]
    \centering
    \includegraphics[width=0.47\textwidth]{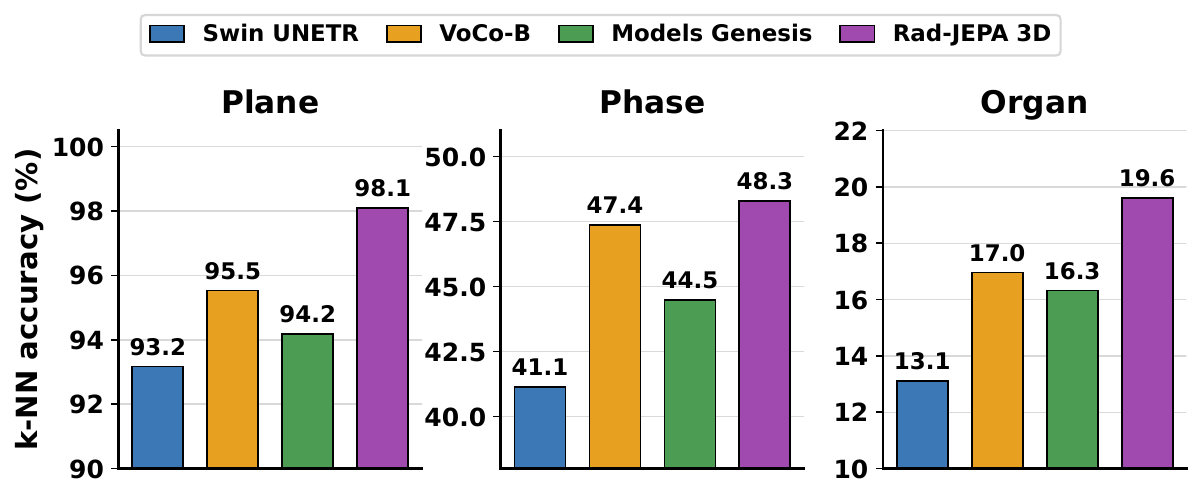}
    \caption{Frozen-encoder $k$-NN probing accuracy on three CT tasks.
    Encoders are self-supervised and kept frozen; a $k$-NN probe classifies their
    representations, isolating representation quality from fine-tuning. Tasks are
    ordered by increasing granularity: scan plane (chance $50.0\%$), contrast
    phase (chance $10.0\%$), and organ (chance $2.9\%$). RadJepa-Hybrid leads on
    all three, with the largest margin on the finest-grained task.}
    \label{fig:knn_probing}
\end{figure}

\label{sec:results}
\begin{table*}[!ht]
\centering
\setlength{\tabcolsep}{4pt}
\renewcommand{\arraystretch}{1.1}
\small
\resizebox{\textwidth}{!}{
\begin{tabular}{l c l l|ccccc|c}
\toprule
\textbf{Method}
& \textbf{Size}
& \textbf{Vision backbone}
& \textbf{Language model}
& \textbf{Plane}
& \textbf{Phase}
& \textbf{Organ}
& \textbf{Abnormality}
& \textbf{Location}
& \textbf{Mean} \\
\midrule
RadFM
& 14.0B
& 3D ViT
& MedLLaMA-13B
& 19.65 & 28.70 & 16.80 & 18.92 & 14.88 & 19.79 \\

M3D-LaMed
& 6.9B
& M3D-CLIP
& LLaMA-2-7B
& 98.80 & 79.75 & 74.75 & 66.65 & 58.94 & 75.78 \\

OmniV-Med-Tiny
& 1.5B
& SigLIP
& Qwen2.5-1.5B
& 98.95 & 91.35 & 75.15 & 66.24 & 60.78 & 78.49 \\

OmniV-Med
& 7.0B
& SigLIP
& Qwen2.5-7B
& 98.75 & \textbf{91.65} & 76.35 & 66.80 & 61.86 & 79.08 \\

Med3DVLM
& 7.6B
& DCFormer-SigLP
& Qwen2.5-7B-Instruct
& 99.15 & 87.50 & 77.45 & 70.17 & 64.49 & 79.75 \\
\midrule
\textbf{Our}
& 3.0B
& Rad-JEPA 3D
& Qwen2.5-3B
& 99.20
& 81.88
& 78.78
& 79.28
& 66.17
& 81.06 \\
\textbf{Our}
& 4.0B
& Rad-JEPA 3D
& Qwen3-4B
& \textbf{99.20}
& 81.20
& \textbf{79.70}
& \textbf{81.00}
& \textbf{67.20}
& \textbf{81.66} \\
\bottomrule
\end{tabular}
}
\caption{
Closed-ended VQA accuracy (\%) on the M3D-VQA benchmark.
The best result in each category is shown in \textbf{bold}. Our VLM with Rad-JEPA 3D as the visual encoder has the small number of parameters but achieves competitive results with state-of-the-art with larger number of parameters.
}
\label{tab:close_ended_vqa}
\end{table*}

\begin{table*}[!ht]
\centering
\setlength{\tabcolsep}{4pt}
\renewcommand{\arraystretch}{1.1}
\small
\resizebox{\textwidth}{!}{
\begin{tabular}{l c l l|cccc}
\toprule
\textbf{Method}
& \textbf{Size}
& \textbf{Vision backbone}
& \textbf{Language model}
& \textbf{BLEU}
& \textbf{ROUGE}
& \textbf{METEOR}
& \textbf{BERTScore} \\
\midrule
RadFM
& 14.0B
& 3D ViT
& MedLLaMA-13B
& 16.39 & 26.13 & 21.33 & 88.72 \\
M3D-LaMed
& 6.9B
& M3D-CLIP
& LLaMA-2-7B
& 49.38 & 52.39 & 33.58 & 91.53 \\
Med3DVLM
& 7.6B
& DCFormer-SigLP
& Qwen2.5-7B-Instruct
& 52.38 & 56.31 & 36.76 & 92.18 \\
\midrule
\textbf{Our}
& 3.0B
& Rad-JEPA 3D
& Qwen2.5-3B
& \underline{80.31}
& \underline{84.54}
& \underline{59.33}
& 90.77 \\
\textbf{Our}
& 4.0B
& Rad-JEPA 3D
& Qwen3-4B
& \textbf{86.16}
& \textbf{84.97}
& \textbf{60.81}
& \textbf{97.61} \\
\bottomrule
\end{tabular}
}
\caption{Open-ended VQA on the M3D-VQA test set: models write a free-text
answer from a CT volume and question, scored against references with
BLEU-1, ROUGE-L, METEOR, and BERTScore F1.
Our 4.0B model leads every metric while using about half the parameters
of the strongest baseline. Best in \textbf{bold}, second-best
\underline{underlined}.}
\label{tab:open_vqa}
\end{table*}
\begin{table*}[!ht]
\centering
\setlength{\tabcolsep}{3.0pt}
\renewcommand{\arraystretch}{1.1}
\small
\resizebox{\textwidth}{!}{
\begin{tabular}{l c l l c c c c|c}
\toprule
\multirow{2}{*}{\textbf{Method}} &
\multirow{2}{*}{\textbf{Size}} &
\multirow{2}{*}{\textbf{Vision backbone}} &
\multirow{2}{*}{\textbf{Language backbone}} &
\multicolumn{4}{c|}{\textbf{MCQ (\%)}} &
\multirow{2}{*}{\textbf{AVG}} \\
\cmidrule(lr){5-8}
& & &
& \textbf{DIR} & \textbf{EXT} & \textbf{VOL} & \textbf{COMP} & \\
\midrule

Med-2E3 
& 3.8B
& M3D-CLIP + SigLIP-L/16
& Phi-3-mini-128K-Instruct
& 34.10 & 37.48 & 45.41 & 46.19 & 41.19 \\

M3D-LaMed 
& 6.9B
& M3D-CLIP
& LLaMA-2-7B
& 0.12 & 27.13 & 33.22 & 13.30 & 21.92 \\

BTB3D
&8.0B
& 3D-CNN
& LLaMA-3.1-8B
& 39.66 & 35.58 & 22.51 & 30.21 & 31.86 \\

Med3DVLM 
& 7.6B
& DCFormer-SigLIP
& Qwen2.5-7B-Instruct
& \textbf{71.10} & 55.23 & 41.37 & 54.65 & 55.59 \\
\midrule
Our
& 3.0B
& Rad-JEPA 3D
& Qwen2.5-3B
& 58.45
& 51.11
& 41.43
& 59.58
& 52.64 \\
\textbf{Our}
& 4.0B
& Rad-JEPA 3D
& Qwen3-4B
& 64.62
& \textbf{58.17}
& \textbf{50.84}
& \textbf{59.02}
& \textbf{58.16} \\

\bottomrule
\end{tabular}
}
\caption{Spatial reasoning benchmark. Accuracy results of MCQ over directional (DIR), extent/size/shape (EXT), volume-magnitude (VOL), and comparative (COMP)
questions. Our 4.0B model has the best average despite being roughly half
the size of the strongest baseline, leading on EXT, VOL, and COMP but
trailing Med3DVLM on DIR. Best in \textbf{bold}.}
\label{tab:spatial}
\end{table*}
\vspace{-2mm}
\subsection{Quantitative results}

\paragraph{Representation learning comparison.}
Figure~\ref{fig:knn_probing} compares Rad-JEPA~3D against prior
self-supervised representation learning methods for 3D medical imaging under
frozen-encoder $k$-NN probing. Rad-JEPA~3D achieves the strongest results, reaching $98.1\%$ on Plane, $48.3\%$ on Phase, and $19.6\%$ on
Organ. Relative to VoCo-B, the strongest baseline on every task, this
corresponds to $+2.6$ points on Plane, $+0.9$ points on Phase, and $+2.6$
points on Organ. The ordering of the baselines is consistent across tasks,
with VoCo-B ahead of Models Genesis and Swin UNETR throughout, suggesting
that contrastive volume-level objectives transfer better than
reconstruction-based ones in this setting; our latent-space prediction
objective improves on both, and its margin is largest on Plane and Organ,
where the label space is respectively coarsest and finest. These results
indicate that predicting in latent space yields representations that remain
discriminative as task granularity increases.
\vspace{-2mm}
\paragraph{Closed-ended VQA comparison.}
Table~\ref{tab:close_ended_vqa} reports closed-ended accuracy on M3D-VQA.
Rad-JEPA~3D reaches the best mean accuracy of $81.66\%$, ahead of Med3DVLM
($79.75\%$) with roughly half the parameters. The largest gain is on
Abnormality detection, $81.00\%$ versus $70.17\%$ ($+10.83$), followed by
Organ recognition at $79.70\%$ ($+2.25$) and Location at $67.20\%$
($+2.71$). These are exactly the categories that require knowing
where a structure sits in the volume, which is what our
self-supervised objective is built to induce: the model trails only on
Phase, a global contrast-timing cue that carries no spatial component.
Both Rad-JEPA~3D variants outperform Med3DVLM in mean accuracy:
the 3.0B model reaches $81.06\%$, while the 4.0B model reaches
$81.66\%$.
\vspace{-1mm}
\paragraph{Open-ended VQA comparison.}
Table~\ref{tab:open_vqa} evaluates free-form answer generation on the
same benchmark. Rad-JEPA~3D leads on all four metrics, reaching $86.16$
BLEU-1 and $60.81$ METEOR against $52.38$ and $36.76$ for Med3DVLM. The
margin is larger here than in the closed-ended setting: when the model must
produce an answer rather than select one, it can no longer fall back on
answer priors, so the quality of the visual representation matters more.
Together with the closed-ended results, this supports our central claim that
spatial and geometric organization in the encoder, not language-model scale,
drives volumetric reasoning.
\vspace{-1mm}
\paragraph{Spatial Reasoning comparison.}
Table~\ref{tab:spatial} evaluates spatial reasoning on the SpatialMed benchmark. Rad-JEPA~3D achieves the best average score of 58.16\%,
outperforming Med3DVLM by $+2.57$. The improvement is consistent across three of the four sub-tasks: our
model leads on EXT ($58.17$ vs.\ $55.23$), VOL ($50.84$ vs.\ $41.37$, a $+9.47$
margin), and COMP ($59.02$ vs.\ $54.65$). The large VOL gain is particularly
telling, as volume-magnitude reasoning requires integrating structure across many axial slices---precisely the cross-plane continuity our hybrid Mamba branch is intended to capture. The exception is DIR, where Med3DVLM leads ($71.10$ vs.\ our $64.62$). These results show that the visual information captured from Rad-Jepa 3D plays a vital role for the improvement of spatial reasoning.
\subsection{Ablation Studies}
\paragraph{The hybrid block and HSOR are complementary, not redundant.}
Table~\ref{tab:ablation} evaluates the hybrid Mamba--GQA block and HSOR using
$k$-NN probing ($k{=}20$) on the Plane, Phase, and Organ tasks. Each component
improves the pure-Mamba baseline on its own, and most visibly on Phase: the
hybrid block raises accuracy from $40.2\%$ to $44.2\%$, and HSOR to $44.4\%$.
Neither, however, improves Organ recognition in isolation ($15.9\%$ and
$16.2\%$, against a $15.9\%$ baseline). Combining the two yields the best
result on every task---$98.1\%$ on Plane, $48.3\%$ on Phase, and $19.6\%$ on
Organ, for a mean of $55.3\%$---including a $+8.1$-point gain on Phase and a
$+3.7$-point gain on Organ over the baseline. The Organ result is the clearest
evidence of complementarity: the combined gain exceeds the sum of the
individual gains, which is what we would expect if the hybrid block enriches
token-level contextual modeling while HSOR reduces redundancy across hidden
dimensions, rather than the two acting on the same source of error.

\begin{table}[!ht]
\centering
\setlength{\tabcolsep}{4pt}
\renewcommand{\arraystretch}{1.15}
\small
\begin{tabular}{cc|cccc}
\toprule
\textbf{Hybrid} & \textbf{HSOR} & \textbf{Plane} & \textbf{Phase} & \textbf{Organ} & \textbf{Mean} \\
\midrule
 &  & 96.0 & 40.2 & 15.9 & 50.7 \\
 & \checkmark & 97.3 & 44.4 & 16.2 & 52.6 \\
\checkmark &  & 96.8 & 44.2 & 15.9 & 52.3 \\
\checkmark & \checkmark & \textbf{98.1} & \textbf{48.3} & \textbf{19.6} & \textbf{55.3} \\
\bottomrule
\end{tabular}
\caption{Impact of our HSOR and the hybrid Mamba-Transformer architecture on representation quality. We report k-NN (k=20) accuracy on three downstream classification tasks from M3D-VQA. HSOR denotes the integration with our HSOR. Hybrid denotes the Mamba-Transformer block with per-token routing.}
\label{tab:ablation}
\end{table}

\vspace{-2mm}
\paragraph{Masking must be aggressive enough to force spatial inference.}
Table~\ref{tab:ablation_mask_ratio} varies the pretraining mask ratio.
Performance is comparatively flat over $0.50$--$0.60$ and $0.85$ (mean
$52.2\%$--$52.9\%$) but rises sharply at $r{=}0.75$, which is best on every
task: $98.1\%$ on Plane, $48.3\%$ on Phase, $19.6\%$ on Organ, and $55.3\%$
mean, $+2.4$ points above the next-best setting. The shape of the curve is
informative rather than incidental: low ratios leave enough of the volume
visible that the pretext task can be solved by local interpolation, weakening
the pressure to infer global spatial structure, whereas at $0.85$ too little
context remains to predict from and the mean falls back to $52.9\%$. The
optimum at $0.75$ shows the high masking ratios found effective in JEPA and reflects the heavy spatial redundancy of volumetric CT---most of a
scan is predictable from a modest visible fraction, so a demanding mask is
what makes the pretext task teach spatial reasoning.

\begin{table}[!ht]
\centering
\begin{tabular}{l|ccccc}
\toprule
\textbf{Mask Ratio} & \textbf{Plane} & \textbf{Phase} & \textbf{Organ} & \textbf{Mean} \\
\midrule
0.50 & 97.1 & 44.3 & 16.9 & 52.8 \\
0.60 & 97.3 & 43.2 & 16.2 & 52.2 \\
\textbf{0.75} & \textbf{98.1} & \textbf{48.3} & \textbf{19.6} & \textbf{55.3} \\
0.85 & 96.6 & 44.2 & 17.8 & 52.9 \\
\bottomrule
\end{tabular}
\caption{Ablation study on mask ratio for V-JEPA Mamba SSL pretraining. All models use random masking, unidirectional scan, patch size $8\times8\times8$, and predictor dim 192 with 4 layers. Best results are in \textbf{bold}.}
\label{tab:ablation_mask_ratio}
\end{table}

\begin{figure}[!ht]
    \centering
    \includegraphics[width=0.4\textwidth]{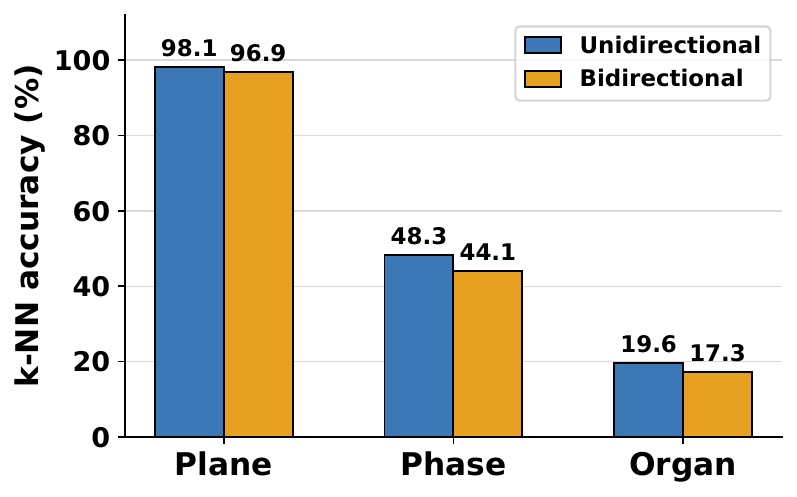}
    \caption{Comparison of unidirectional and bidirectional Mamba
scanning. Unidirectional scanning surpasses on all three tasks
while requiring only one Mamba scan direction.}
    \label{fig:scan_comp}
\end{figure}

\vspace{-2mm}
\paragraph{A single forward scan already captures volumetric context.}
Figure~\ref{fig:scan_comp} compares unidirectional and bidirectional Mamba
scanning. Unidirectional scanning is better on all three tasks and on mean
accuracy ($55.3$ vs.\ $52.8$), with the largest gap on Phase
($48.3$ vs.\ $44.1$), while the bidirectional variant doubles scan cost.
The reason is that, unlike causal language modeling---where a
backward pass supplies new information---our tokens are jointly
available, and the predictive target is bidirectional by construction: each
cube is already reconstructed from full surrounding context through the JEPA
objective and the GQA branch's all-to-all attention. A reverse Mamba scan is
therefore largely redundant, and we default to the cheaper unidirectional
variant.

\vspace{-2mm}
\paragraph{Cube size trades axial resolution against anatomical context.}
Table~\ref{tab:ablation_patch_size} studies the tokenizer's cube size, which
sets the granularity of the token sequence. The largest cubes
($16\times16\times16$) coarsen the representation past useful resolution and
are worst overall ($51.6\%$ mean), while the thinnest axial cubes
($4\times16\times16$) match the best Plane accuracy ($98.1\%$) but fragment
cross-slice anatomy, leaving Phase and Organ at $42.6\%$ and $16.5\%$.
The $8\times8\times8$ configuration ties for the best Plane
accuracy and achieves the best Phase and Organ accuracies, resulting
in the highest mean accuracy of $55.3\%$, and we adopt it as our default.
The comparison with $8\times16\times16$ is instructive: holding axial depth
fixed at $8$ and halving the in-plane extent gains $+4.5$ points on Phase and
$+1.1$ on Organ, indicating that in-plane granularity, not only axial depth,
limits the tasks that depend on fine anatomical detail. Cube size is therefore
not a neutral efficiency knob but a spatial-resolution choice, and
organ-level tasks are the most sensitive to getting it right.

\begin{table}[!ht]
\centering
\begin{tabular}{lccccc}
\toprule
\textbf{Block Size} & \textbf{Plane} & \textbf{Phase} & \textbf{Organ} & \textbf{Mean} \\
\midrule
$4\times16\times16$ & 98.1 & 42.6 & 16.5 & 52.4 \\
$8\times16\times16$ & 97.7 & 43.8 & 18.5 & \textbf{53.3} \\
$8\times8\times8$ & \textbf{98.1} & \textbf{48.3} & \textbf{19.6} & \textbf{55.3} \\
$16\times16\times16$ & 96.4 & 42.7 & 15.7 & 51.6 \\
\bottomrule
\end{tabular}
\caption{Ablation study on patch size for Rad-Jepa pretraining. All models use random masking, mask ratio 0.75, unidirectional scan, and predictor dim 192 with 4 layers. Best results are in \textbf{bold}.}
\label{tab:ablation_patch_size}
\vspace{-2mm}
\end{table}

\vspace{-2mm}
\section{Conclusion}
We presented Rad-JEPA~3D, a joint-embedding predictive framework for
Self-supervised 3D CT representation learning. Its hybrid H-Mamba encoder fuses a Mamba branch with a GQA branch via a per-token
router to capture both inter-slice continuity and cross-plane spatial context.
Moreover, we propose HSOR to improve intermediate representation quality through layer-wise student–teacher alignment, feature decorrelation, and weight orthogonality. Together with the H-Mamba encoder, HSOR yields more discriminative volumetric features, leading to competitive performance on spatially demanding downstream tasks. Pretrained on approximately 120{,}000 CT
scans, Rad-JEPA~3D, corporating with small LLM, attains competitive results with state-of-the-art on several VQA benchmarks with only $4.0$B parameters, which shows that spatial and geometric organization, rather than raw scale, drives volumetric reasoning.

\bibliographystyle{plainnat}
\bibliography{aaai}

\end{document}